\documentclass{article}
\usepackage{spconf,amsmath,graphicx,amssymb,amsthm,amsmath,mathtools}

\usepackage{subfigure,epstopdf,algorithm}
\usepackage{array}
\usepackage{xcolor}
\usepackage{thmtools}

\declaretheorem{proposition}

\usepackage{times}
\usepackage{eqparbox}

\usepackage[noend]{algpseudocode}

\def\K{{\bf K}}
\def\I{{\bf I}}

\def\v{\mbox{vec}}
\title{Multi-kernel regression for graph signal processing}
\name{Arun~Venkitaraman, Saikat Chatterjee, Peter H{\"a}ndel	\thanks{The authors would like to acknowledge the support received from the Swedish Research Council.}
}
	\address{
	Department of Information Science and Engineering,                    
	School of Electrical Engineering\\            
	KTH Royal Institute of Technology,  
	SE-100 44 Stockholm, Sweden                 \\
	arunv@kth.se, sach@kth.se, ph@kth.se
}

\begin{document}
\ninept
\maketitle
\begin{abstract}
We develop a multi-kernel based regression method for graph signal processing where the target signal is assumed to be smooth over a graph. In multi-kernel regression, an effective kernel function is expressed as a linear combination of many basis kernel functions. We estimate the linear weights to learn the effective kernel function by appropriate regularization based on graph smoothness. We show that the resulting optimization problem is shown to be convex and propose an accelerated projected gradient descent based solution. Simulation results using real-world graph signals show efficiency of the multi-kernel based approach over a standard kernel based approach.
\end{abstract}
\section{Introduction}
\label{sec:intro}


Kernel regression is a nonlinear learning approach used extensively in classification, regression, and clustering problems \cite{Saunders98ridgeregression, Bishop, kernel_deeplearning}. Typically, kernel regression needs training with a significant amount of reliable data to ensure a good prediction performance. In the context of limited and noisy data availability, we recently proposed kernel regression for target signals by incorporating an additional structure that the target signal is smooth over a graph \cite{Arun_kergraph}. 
There are prior works in the literature on using graph structures for kernel regression \cite{Smola2003, diffusionkernels, kergraph1,kergraph3}. These existing works do not use the direct approach of predicting the complete target signal by enforcing smoothness constraint over a graph. The prior works consider kernels defined among nodes of the same graph signal for signal completion or recovery.

Performance of kernel-based approaches varies significantly with the choice of the kernel functions and parameters. The optimal choice of the kernel parameter is a difficult problem to solve analytically or through cross-validation. A natural approach is to express a general kernel function that is a linear combination of many known 'basis' kernels and find their optimal combination adapted to the given data \cite{multikernel_7}. This is referred to as multi-kernel learning in the literature and has been shown to be useful in many applications \cite{multikernel_3,multikernel_4,multikernel_5,multikernel_6}. In this article, our contribution is to develop a multi-kernel learning approach for graph signal processing where we predict vector targets. We assume that the vector target signals that we predict are smooth graph signals over an underlying graph. The resulting multi-kernel learning approach is shown to be a convex optimization problem. We then develop a projected gradient descent algorithm to learn the multi-kernel parameters by using Nesterov's method \cite{Nesterov}. 
There exist recent works on developing multi-kernel regression for graph signals in \cite{multikernel_1, multikernel_2}, where the prediction is made on a subset of nodes of a graph from the signals at the remaining nodes through a kernel regression across nodes. Our proposed approach is different from these works by formulation. We explicitly consider that the target is a vector signal lying over a graph, and the input signal is not necessarily over a graph. Our goal is then to make prediction for the smooth graph signal target given a new input signal by learning the parameters of multi-kernel function.

\subsection{Kernel regression for graph signals}
\label{sec:format}
We provide background in this section on recently proposed kernel regression for graph signal processing \cite{multikernel_2}.
In graph signal processing, let $\mathcal{G}=(\mathcal{V},\mathcal{E},\mathbf{A})$ denotes a graph with $M$ vertices indexed by set $\mathcal{V}$, edge set $\mathcal{E}$, and the adjacency matrix $\mathbf{A}=[a_{ij}]$, $a_{ij}>0$. A graph signal over $\mathcal{G}$ is a vector in $\mathbb{R}^M$ whose components denote the values of the signal at the nodes indexed by $\mathcal{V}$ \cite{Shuman,Sandry1}. We consider only undirected graphs in our analysis, which corrsponds to symmetric $\mathbf{A}$. The graph-Laplacian matrix $\mathbf{L}$ of $\mathcal{G}$ is then defined as \cite{Chung}
%
$\mathbf{L}=\mathbf{D}-\mathbf{A}$,
where $\mathbf{D}=\mathrm{diag}(d_1,d_2\cdots,d_M)$ denotes the diagonal degree matrix with $d_{i}=\sum_ja_{ij}$. By construction, $\mathbf{L}$ is symmetric and positive semi-definite. The smoothness of a graph signal $\mathbf{y}=[y(1),y(2),\cdots,y(M)]^\top$ is quantified by the following quadratic form:
\begin{equation*}
l(\mathbf{y})=\mathbf{y}^\top\mathbf{L}\mathbf{y}=\sum_{(ij)\in\mathcal{E}}a_{ij}(y(i)-y(j))^2.
\end{equation*}
A small $l(\mathbf{y)}$ implies that $\mathbf{y}$ is smooth since its value varies smoothly over connected nodes.

We next briefy discuss a standard kernel regression. 
Consider a set of $N$ training observations of input $\mathbf{x}_n\in\mathbb{R}^L$ and the corresponding target $\mathbf{t}_n\in\mathbb{R}^M$. Let $\pmb\phi(\mathbf{x})\in\mathbb{R}^K$ denotes a vector function of the input $\mathbf{x}$. Then, kernel regression makes prediction of output $\mathbf{y}$ for a new input $\mathbf{x}$ of the form \cite{Bishop}
\begin{equation*}
\mathbf{y}=\mathbf{W}^\top\pmb\phi(\mathbf{x}),
\end{equation*}
where the regression coefficient matrix is $\mathbf{W}$, found by
	\begin{equation}
\label{eq:cost}
\mathbf{W}=\arg\min_{\mathbf{W}}\sum_{n=1}^N \|\mathbf{t}_n-\mathbf{y}_n\|_2^2 + \alpha \, \mbox{tr}(\mathbf{W}^\top\mathbf{W}),\,\,\alpha\geq0.
\end{equation}
 The optimal $\mathbf{W}$ depends on inner products of form $\pmb\phi(\mathbf{x})^\top\pmb\phi(\mathbf{x}')$ \cite{Bishop}. On generalizing the inner product to a kernel function  $k(\mathbf{x},\mathbf{x}')$, the predicted output takes the form:
\begin{equation}
\mathbf{y}=\mathbf{\Psi}^\top\mathbf{k}(\mathbf{x}),\nonumber
\end{equation}
where $\pmb\Psi=(\mathbf{K}+\alpha\mathbf{I})^{-1}\mathbf{T}$, $\mathbf{K}\in\mathbb{R}^{N\times N}$ denotes the kernel matrix with $(m,n)$th entry is equal to $k(\mathbf{x}_{m},\mathbf{x}_n)$, $\mathbf{T}=[\mathbf{t}_1\,\mathbf{t}_2\cdots \mathbf{t}_N]^\top$ is the target matrix, and $\mathbf{k}(\mathbf{x})=[k(\mathbf{x}_{1},\mathbf{x}),k(\mathbf{x}_{2},\mathbf{x}),\cdots, k(\mathbf{x}_{N},\mathbf{x})]^\top$.

We finally discuss kernel regression for graph signal processing, proposed in \cite{Arun_kergraph}. It is assumed that the entire target signal is smooth over the underlying graph $\mathcal{G}$. In kernel regression for graph signal processing, the following optimization problem is solved \cite{Arun_kergraph}:
\begin{equation}
\label{eq:cost_krg}
\min_{\mathbf{W}}\sum_{n=1}^N \|\mathbf{t}_n-\mathbf{y}_n\|_2^2 + \alpha \, \mbox{tr}(\mathbf{W}^\top\mathbf{W})+\beta \sum_{n=1}^Nl( \mathbf{y}_n),\,\,\alpha,\beta\geq0,
\end{equation}
where the predicted signal $\mathbf{y}_n=\mathbf{W}^\top\pmb\phi(\mathbf{x}_n)$ is enforced to be smooth over $\mathcal{G}$ by using the regularization term $\sum_{n=1}^N l({\mathbf{y}_n})$. The optimization problem \eqref{eq:cost_krg} is equivalent to the following optimization problem \cite{Arun_kergraph}
\begin{equation} 
\label{eq: krg_cost}
\begin{array}{r}
\min_{\pmb\Psi}( \mbox{tr}(\mathbf{T}^\top\mathbf{T})-2\mbox{tr}\left(\mathbf{T}^\top\bf{K}\mathbf{\Psi} \right) +\mbox{tr}\Large( \mathbf{\Psi}^\top\bf{K}\bf{K}\mathbf{\Psi}) \\
\quad+\alpha\, \mbox{tr}(\mathbf{\Psi}^\top\bf{K}\mathbf{\Psi}) +  \beta \,\mbox{tr}\left( \mathbf{\Psi}^\top\bf{K}\bf{K}\mathbf{\Psi}\mathbf{ L}\right)).
\end{array}
\end{equation}
The optimal prediction $\mathbf{y}=\pmb\Psi^\top\mathbf{k}(\mathbf{x})$ is associated with the optimal $\pmb\Psi$ matrix with the following form
\begin{align}
\label{eq: krg_optimal_psi}
\mbox{vec}(\mathbf{\Psi})=\left[(\mathbf{I}_M\otimes (\mathbf{K}+\alpha\mathbf{I}_N))+(\beta \mathbf{L}\otimes \mathbf{K})\right]^{-1}\mbox{vec}(\mathbf{T}),
\end{align}
where $\mathrm{vec}(\cdot)$ and $\otimes$ denote the vectorization and Kronecker product operations, respectively. We note that, if there is no graph smoothness penalty ($\beta = 0$) then \eqref{eq:cost_krg} reduces to standard kernel regression.

\section{Multi-kernel regression for graph signals}
We now develop multi-kernel regression for graph signal processing. In multi-kernel regression we use a general kernel that is a linear combination of multiple pre-specified kernels, as follows
\begin{equation} 
k(\mathbf{x},\mathbf{x}')=\sum_{s=1}^S\rho_s k_s(\mathbf{x},\mathbf{x}'),
\label{eq:multi_kernel_form}
\end{equation}
where $k_s(\mathbf{x},\mathbf{x}')$ denotes the $s$'th pre-specified kernel function. The weight vector $\pmb\rho=[\rho_1,\rho_2,\cdots,\rho_S]$ is unknown.  We assume that $\pmb\rho$ has a bounded q-norm $\|\pmb\rho\|_q$, where $q$ may be 1 or 2.
Let us use $\mathbf{K}_s$ to denote the kernel matrix associated with the $s$'th kernel function $k_s(\mathbf{x}_{m},\mathbf{x}_n)$. We then have the general kernel matrix as $\mathbf{K}=\sum_{i=s}^S\rho_s\mathbf{K}_s$. Now, generalizing the kernel regression for graph signal processing optimization problem \eqref{eq: krg_cost}, we learn the optimal weight vector and regression coefficients by the following optimization problem:
\begin{equation}
\begin{array}{r}
\label{eq:cost_multikernel}
 \min_{\pmb\rho}\min_{\pmb\Psi} -2\mbox{tr}\left(\mathbf{T}^\top\bf{K}\mathbf{\Psi} \right) +\mbox{tr}\left( \mathbf{\Psi}^\top\bf{K}\bf{K}\mathbf{\Psi}\right) \\
+\alpha\, \mbox{tr}(\mathbf{\Psi}^\top\bf{K}\mathbf{\Psi}) +  \beta \,\mbox{tr}\left( \mathbf{\Psi}^\top\bf{K}\bf{K}\mathbf{\Psi}\mathbf{ L}\right),\,\,\alpha,\beta\geq0\\
\mbox{subject to  }\pmb\rho\succeq\mathbf{0}, \|\pmb\rho\|_q\leq R
,\,\,\mathbf{K}=\sum_{i=s}^S\rho_s\mathbf{K}_s
\end{array}
\end{equation}
where $\|\pmb\rho\|_q\leq R$ is a regularization constraint. We note that for a fixed $\pmb\rho$, the general kernel function is known. Therefore, for a given $\pmb\rho$, the optimization problem becomes same as the problem in \eqref{eq: krg_optimal_psi}. In order to solve \eqref{eq:cost_multikernel}, we thus substitute $\pmb\Psi$ for a given $\pmb\rho$ and simplify the problem in the following equivalent form
\begin{eqnarray}
\label{eq:cost_multikernel_2}
\min_{\pmb\rho}\mbox{vec}(\mathbf{T})^\top \mathbf{B}(\pmb\rho)\mbox{vec}(\mathbf{T})
\mbox{  subject to  }\pmb\rho\succeq\mathbf{0}, \|\pmb\rho\|_q\leq R,
\end{eqnarray}
where
\begin{equation}
\label{B_multi_kernel}
\mathbf{B}(\pmb\rho)=-(\mathbf{I}_M\otimes\mathbf{K})\left[\mathbf{I}_M\otimes (\mathbf{K}+\alpha\mathbf{I}_N)+\beta\mathbf{L}\otimes \mathbf{K}\right]^{-1}.
\end{equation}
The steps in the simplification of \eqref{eq:cost_multikernel} to \eqref{eq:cost_multikernel_2} is shown in the appendix.

\subsection{Theoretical analysis}
\begin{proposition}
The optimization problem \eqref{eq:cost_multikernel_2} is convex in $\pmb\rho$.
\end{proposition}
\begin{proposition}
Let us define $\gamma(\pmb\rho)\triangleq\mbox{vec}(\mathbf{T})^\top \mathbf{B}(\pmb\rho)\mbox{vec}(\mathbf{T})$. We state that $\gamma(\pmb\rho)$ is a decreasing function of $\pmb\rho $ and takes the maximum value of $0$ at $\pmb\rho=\mathbf{0}$.
\end{proposition}
\noindent \emph{Proof:} Proofs of propositions 1 and 2 are provided in the appendix.
\begin{proposition}
	The solution of \eqref{eq:cost_multikernel_2} satisfies the boundary condition $\|\pmb\rho\|_q=R$.
	\begin{proof}
		This follows from Proposition 2 that $\gamma(\pmb\rho)$ is a decreasing function for $\pmb\rho\succeq\mathbf{0}$ and takes the maximum value at $\mathbf{0}$.
	\end{proof}
\end{proposition}
\begin{proposition}
	The gradient $\nabla\gamma(\pmb\rho) $ of $\gamma(\pmb\rho)$ is given by $\nabla\gamma(\pmb\rho)=[\nabla\gamma_1,\nabla\gamma_2,\cdots,\nabla\gamma_S]$ where
	\begin{eqnarray}
	\nabla\gamma_s&=&-\mathrm{vec}(\pmb\Psi)^\top(\mathbf{I}_M\otimes \mathbf{K}_s)\mathrm{vec}(\pmb\Psi),\nonumber\\
		 \mathrm{vec}(\pmb\Psi)&=&\left[(\mathbf{I}_M\otimes (\mathbf{K}+\alpha\mathbf{I}_N))+(\beta \mathbf{L}\otimes \mathbf{K})\right]^{-1}\mathrm{vec}(\mathbf{T})\nonumber
		 \end{eqnarray}
	 Also, $\nabla\gamma(\pmb\rho)\preceq\mathbf{0}$.
	 \begin{proof}
	 	The expression for the derivative is obtained using standard matrix derivative relations and the proof is omitted here for brevity. Each kernel matrix is positive semidefinite by definition, and hence $\mathbf{I}_M\otimes \mathbf{K}_s\succeq0$, $\forall s$. Since each component of the gradient is a negative quadratic form with a positive semidefinite matrix, $\nabla\gamma_s\leq0$ $\forall s$. Hence, $\nabla\gamma(\pmb\rho)\preceq\bf{0}$. 
	 \end{proof}
\end{proposition} 

\subsection{Estimating kernel parameters}
We have shown that the optimization problem in \eqref{eq:cost_multikernel_2} is convex. However, we find that it is non-trivial to express the optimization problem into a standard optimization form such that a convex optimization toolbox, like CVX \cite{cvx}, can be used. Hence, we use a projected gradient descent approach to solve the optimization problem. This is possible due to the property that $\gamma(\pmb\rho)$ has a gradient in closed-form (see Proposition 4). In order to improve the rate of convergence over that of the projected gradient descent, we use Nesterov's accelerated gradient approach \cite{Nesterov}. For successive iterations of the gradient descent, we use a step-size inversely proportional to the number of iterations. The resulting steps of the gradient decent search are described in Algorithm \ref{multi_kernel_algo}.
	\begin{algorithm}[t]
		\label{multi_kernel_algo}
	\caption{ Multi-kernel regression for graph signal processing}
	\begin{algorithmic}[1]
		\State Initialize $\mu(0)$, $\pmb\rho(0)=0$, $\lambda(0)=1$ 
		\State $i_{max}$ is the maximum number of iterations
		\While {$i< i_{max}$ \mbox{OR} $\|\pmb\rho(i)-\pmb\rho(i-1)\|^2_2>\epsilon$} 
			\State $\mu(i)=\mu(0)/i$           \State$\mathbf{s}(i)=\pmb\rho(i-1)-\mu(i)\nabla\gamma(\pmb\rho(i-1))$
			\State
			$\mathbf{z}(i)=\arg\min_{z}\|\mathbf{s}(i)-\mathbf{z}\|_2^2$,\newline
			$\mbox{subject to }\,\,\|\mathbf{z}\|_1\leq R, \,\mathbf{z}\succeq 0$
			\State Set $\nu(i)=\frac{\lambda(i-1)-1}{\lambda(i+1)}$,
			$\lambda(i)=\frac{1+\sqrt{1+4\lambda(i-1)^2}}{2}$ \hfill (Nesterov)
		\State $\pmb\rho(i)=(1-\nu(i))\mathbf{z}(i)+\nu(i)\pmb\rho(i-1)$ \hfill (Nesterov)
		\State $i\to i+1$
		\EndWhile
	\end{algorithmic}
\end{algorithm}

\section{Experiments}

\begin{table}[t]
	\centering
	\caption{List of regularization parameter values}
	\begin{tabular}{|c|c|c|c|c|} 
		\hline
		$N$ & Linear  &Kernel regression & $R$ & $\mu(0)$ \\
		&regression & on graph ($\sigma=1$) & &\\
		&$(\alpha)$& $(\alpha,\beta)$& &\\
		\hline
		4& 4.3  &(0.02,5.5) &  5&0.01\\
		8&4.3   &(0.03,5.5)&  5&0.01\\
		16&4.3  &(0.06,5.5) &  5&0.01\\
		30&4.3  &(0.1,5.5) &  5&0.01\\
		\hline
	\end{tabular}
\label{table:multikernel_table}
\end{table}

\begin{figure}[t]
	\centering
	\includegraphics[width=3.2in]{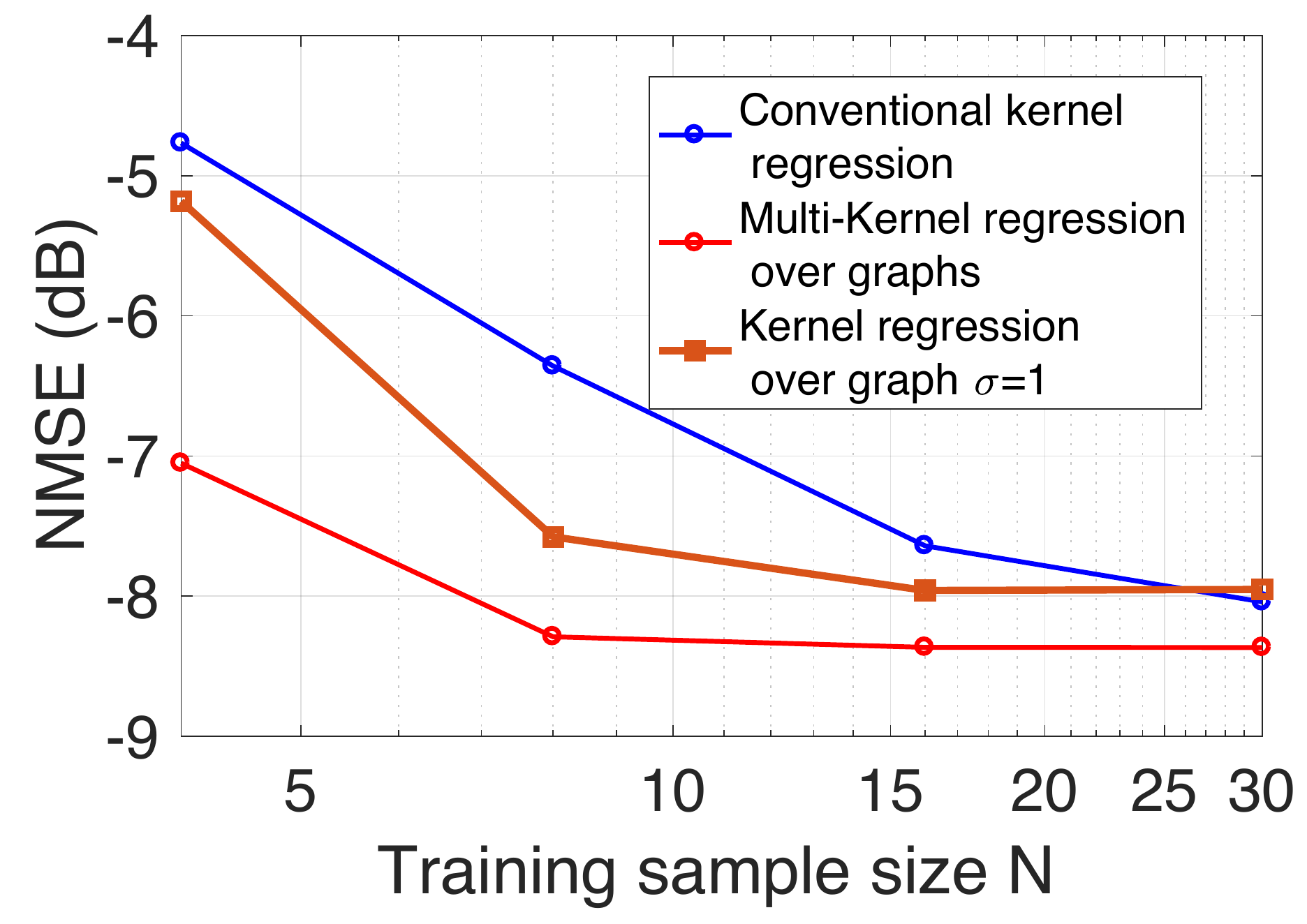}
	\caption{NMSE as function of training sample size at SNR  of 0dB.}
		\label{fig:multi_ker_temp_1}
	\vspace{-.in}
\end{figure}

\begin{figure}[t]
	\centering
	\includegraphics[width=3.2in]{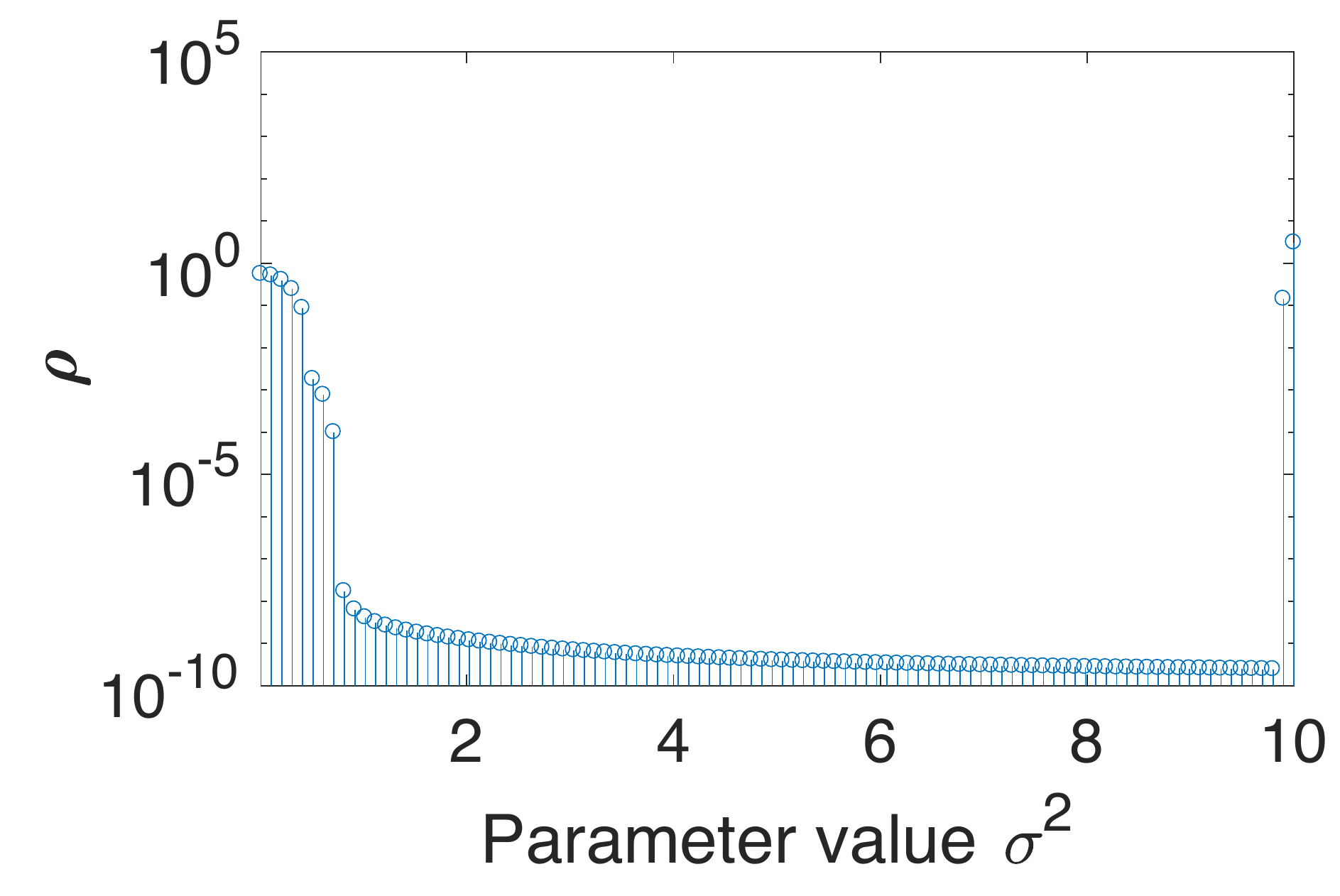}
	\caption{An instance of $\pmb\rho$ for the choice $q=1$. }
	\vspace{-.in}
	\label{fig:multi_ker_temp_2}
\end{figure}

We apply our approach for target prediction on daily temperature measurements over 45 largest cities in Sweden \footnote{The codes used for experiments may be found at:\\
	https://www.kth.se/en/ees/omskolan/organisation/avdelningar/information-science-and-engineering/research/reproducibleresearch}. At a particular time instant, temperature data over cities of a country is a graph signal. We consider 60 temperature measurements during the time period of October to December 2015 \cite{SMHI}. Let $\mathbf{t}_{o,n}\in\mathbb{R}^{45}$ is the vector that contains temperatures of 45 cities at a particular day. The corresponding input vector $\mathbf{x}_n\in\mathbb{R}^{45}$ is true temperature values from the previous day. Our interest is to predict the temperature of cities for the next day from the readings of current day. For the sake of experiments, we assume that in many applications the true target values are unknown or can not be observed. We only have access to noisy targets. To emulate such kind of situations, we deliberately corrupt true target temperature signal $\mathbf{t}_{o,n}$ and use the corrupted (noisy) targets for training the learning algorithms. Then, at the time of testing, we try to predict the true target and check the robustness of learning algorithms. In this experiment, we generated noisy targets $\mathbf{t}_n$ as follows:
\begin{equation}
\mathbf{t}_n=\mathbf{t}_{o,n}+\mathbf{e}_n,\nonumber
\end{equation}
where $\mathbf{e}_n$ denotes the white Gaussian noise at a signal-to-noise ratio (SNR) of $0$ dB.
Let $d_{ij}$ denote the geodesic distance between $i$'th and $j$'th cities. We construct the adjacency matrix of the graph by setting 
\begin{equation}
a_{ij}=\exp{\left(-\frac{d_{ij}^2}{\sum_{i,j}d_{ij}^2}\right)}.\nonumber
\end{equation}
We randomly partition the total dataset into two subsets, each subset contains $30$ samples. Then we use one dataset for training and the other for testing. We consider $S=100$ Gaussian kernels where the parameters (variances) are picked in uniform step size from the span $[0.01,10]$. Then we predict $\mathbf{t}_{o,n}$ using the output $\mathbf{y}_n$ and evaluate normalized mean-squared error (NMSE) performance.
The NMSE is obtained by averaging over $100$ different noise realizations and partitions of the data. We find that  Algorithm 1 typically converges after around 50 steps with $\epsilon=10^{-4}$. We compare the NMSE of our approach with that of the conventional regression using linear kernel $\pmb\phi(\mathbf{x})=\mathbf{x}$, and kernel regression over graphs \cite{Arun_kergraph} with Gaussian kernel with parameter equal to 1. For the conventional regression and kernel regression with single Gaussian kernel, $\alpha$ and $\beta$ resulting in the smallest NMSE for training data are found by exhaustive grid search. For multi-kernel regression, we set $\alpha$ and $\beta$ to be the same as those obtained for the single Gaussian kernel. We consider the case of $q=1$ in our experiments in this paper. We also performed experiments with $q=2$, though they are not reported here for brevity. $R=5$ is experimentally found to be a good choice for all training sample sizes. The parameters used in the experiments are shown in Table~\ref{table:multikernel_table}. The NMSE for testing data for different approaches is shown in Figure~\ref{fig:multi_ker_temp_1}. We observe that the multi-kernel approach clearly outperforms the other two approaches particularly at low training sample sizes. An instance of $\pmb\rho$ learned for a random data partition is shown in Figure~\ref{fig:multi_ker_temp_2}, which demonstrates how the kernels that best explain the data are selected by the algorithm, since we have used $q=1$ which is known to promote sparsity.

\vspace{-.0in}
\section{Concusions}
We proposed a multi-kernel regression for targets that are smooth over a given graph. This was built on our earlier work of kernel regression for smooth signals over graphs which was shown to outperform conventional kernel regression for small training sample sizes and under noisy training. Multi-kernel regression was shown to be a convex optimization problem and some of its properties were studied. We then proposed an accelerated projected descent for evaluating the optimal kernel coefficients. Experiments on real-data demonstrated the potential of our approach for small training sizes and low SNR levels.
\vspace{-.0in}
\section{Appendix}
We derive the steps for simplification of \eqref{eq:cost_multikernel} to \eqref{eq:cost_multikernel_2}.
We know that for fixed $\pmb\rho$ the optimal $\pmb\Psi$ is given by \eqref{eq: krg_optimal_psi}:
\begin{align}
\mbox{vec}(\mathbf{\Psi})=\left[(\mathbf{I}_M\otimes (\mathbf{K}+\alpha\mathbf{I}_N))+(\beta \mathbf{L}\otimes \mathbf{K})\right]^{-1}\mbox{vec}(\mathbf{T}),
\end{align}
Let $\mathbf{B}=\mathbf{I}_M\otimes (\mathbf{K}+\alpha\mathbf{I}_N)$ and $\mathbf{C}=(\beta \mathbf{L}\otimes \mathbf{K})$. Then, $\pmb\Psi=(\mathbf{B}+\mathbf{C})^{-1}\mbox{vec}(\mathbf{T})$.
On using the property that $\mbox{tr}(\mathbf{a}_1^\top\mathbf{a}_2)=\v(\mathbf{a}_1)^\top\v(\mathbf{ a}_2)$ and substituting optimal $\pmb\Psi$ for a fixed $\pmb\rho$, the objective function is given by
\begin{align}
C&=-2\mbox{tr}\left(\mathbf{T}^\top\bf{K}\mathbf{\Psi} \right) +\mbox{tr}\left( \mathbf{\Psi}^\top\bf{K}\bf{K}\mathbf{\Psi}\right) \nonumber\\
&\quad+\alpha\, \mbox{tr}(\mathbf{\Psi}^\top\bf{K}\mathbf{\Psi}) +  \beta \,\mbox{tr}\left( \mathbf{\Psi}^\top\bf{K}\bf{K}\mathbf{\Psi}\mathbf{ L}\right)\nonumber\\
&= -2\mbox{vec}(\mathbf{T})^\top\mbox{vec}(\bf{K}\mathbf{\Psi} ) +\mbox{vec} \mathbf{\Psi})^\top\mbox{vec}(\bf{K}\bf{K}\mathbf{\Psi}) \nonumber\\
&\quad+\alpha\, \mbox{vec}(\mathbf{\Psi})^\top\mbox{vec}(\bf{K}\mathbf{\Psi}) +  \beta \,\mbox{vec}( \mathbf{\Psi})^\top\mbox{vec}(\bf{K}\bf{K}\mathbf{\Psi}\mathbf{ L})\nonumber\\
&= -2\mbox{vec}(\mathbf{T})^\top(\mathbf{I}\otimes\K)\mbox{vec}(\mathbf{\Psi} ) +\mbox{vec} \mathbf{\Psi})^\top(\mathbf{I}\otimes\bf{K}\bf{K})\v(\mathbf{\Psi}) \nonumber\\
&\quad+\alpha\, \mbox{vec}(\mathbf{\Psi})^\top(\I\otimes\bf{K})\v(\mathbf{\Psi}) +  \beta \,\mbox{vec}( \mathbf{\Psi})^\top(\mathbf{L}\otimes\bf{K}\bf{K})\v(\mathbf{\Psi})\nonumber\\
&= -2\mbox{vec}(\mathbf{T})^\top(\mathbf{I}\otimes\K)(\mathbf{B}+\mathbf{C})^{-1}\mbox{vec}(\mathbf{T})\nonumber\\
& \quad+\mbox{vec}(\mathbf{T})^\top(\mathbf{B}+\mathbf{C})^{-1}(\mathbf{I}\otimes\bf{K}\bf{K})(\mathbf{B}+\mathbf{C})^{-1}\mbox{vec}(\mathbf{T}) \nonumber\\
&\quad+\alpha\, \mbox{vec}(\mathbf{T})^\top(\mathbf{B}+\mathbf{C})^{-1}(\I\otimes\bf{K})(\mathbf{B}+\mathbf{C})^{-1}\mbox{vec}(\mathbf{T})\nonumber\\
&\quad+\beta \,\mbox{vec}(\mathbf{T})^\top(\mathbf{B}+\mathbf{C})^{-1}(\mathbf{L}\otimes\bf{K}\bf{K})(\mathbf{B}+\mathbf{C})^{-1}\mbox{vec}(\mathbf{T})\nonumber\\
&= \mbox{vec}(\mathbf{T})^\top[-2(\mathbf{I}\otimes\K)+(\mathbf{B}+\mathbf{C})^{-1}\mathbf{H}](\mathbf{B}\nonumber+\mathbf{C})^{-1}\mbox{vec}(\mathbf{T}).
\end{align}
where $\mathbf{H}=\left[(\mathbf{I}\otimes(\bf{I}+\alpha\K)\K)+ \beta\,(\mathbf{L}\otimes\bf{K}\bf{K})\right]$, and we have used properties of the $\mbox{tr}(\cdot)$ and $\mbox{vec}(\cdot)$ operators in the different steps.
Using distributivity of Kronecker product, we have that 
%
\begin{align}
\mathbf{H}=\left[(\mathbf{I}\otimes(\bf{I}+\alpha\K)\K)
+ \beta \,(\mathbf{L}\otimes\bf{K}\bf{K})\right]=(\mathbf{B}+\mathbf{C})(\I\otimes\K).\nonumber
\end{align} 
Substituting $\mathbf{H}$ in $C$, we get that
\begin{align}
C&=\mbox{vec}(\mathbf{T})^\top[-2(\mathbf{I}\otimes\K)+(\mathbf{I}\otimes\K)](\mathbf{B}\nonumber+\mathbf{C})^{-1}\mbox{vec}(\mathbf{T}).\nonumber\\
&=-\mbox{vec}(\mathbf{T})^\top(\mathbf{I}\otimes\K)(\mathbf{B}+\mathbf{C})^{-1}\mbox{vec}(\mathbf{T}).\nonumber
\end{align}
	\begin{proof}[Proposition 1]
	We prove this by showing that the equivalent problem in \eqref{eq:cost_multikernel} is convex.
	We note that the objective function of \eqref{eq:cost_multikernel}  is the minimum of functions of the form
	\begin{eqnarray*}
		\xi(\pmb\rho)&=-2\mbox{tr}\left(\mathbf{T}^\top\bf{K}\mathbf{\Psi} \right) +\mbox{tr}\left( \mathbf{\Psi}^\top\bf{K}\bf{K}\mathbf{\Psi}\right)\nonumber\\
		& +\alpha\, \mbox{tr}(\mathbf{\Psi}^\top\bf{K}\mathbf{\Psi}) +  \beta \,\mbox{tr}\left( \mathbf{\Psi}^\top\bf{K}\bf{K}\mathbf{\Psi}\mathbf{ L}\right)
	\end{eqnarray*}
	Since minimum of convex functions is also convex, it suffices to prove that $\xi(\pmb\rho)$ is a convex function of $\pmb\rho$. We have that
	\begin{align}
	&\xi(\pmb\rho)=-2\mbox{tr}\left(\mathbf{T}^\top\bf{K}\mathbf{\Psi} \right) +\mbox{tr}\left( \mathbf{\Psi}^\top\bf{K}\bf{K}\mathbf{\Psi}\right)\nonumber\\
	&	+\alpha\, \mbox{tr}(\mathbf{\Psi}^\top\bf{K}\mathbf{\Psi}) +  \beta \,\mbox{tr}\left( \mathbf{\Psi}^\top\bf{K}\bf{K}\mathbf{\Psi}\mathbf{ L}\right)\nonumber\\
	&=-2\mbox{tr}\left(\mathbf{T}^\top\sum_{s=1}^S\rho_s\mathbf{K}_s\large\mathbf{\Psi} \right)\nonumber\\ 
	&+\mbox{tr}\left( \mathbf{\Psi}^\top\sum_{r=1}^S\sum_{s=1}^S\rho_r\rho_s\mathbf{K}_r\mathbf{K}_s\large\mathbf{\Psi}\right)+\alpha\, \mbox{tr}\left(\mathbf{\Psi}^\top\sum_{s=1}^S\rho_s\mathbf{K}_s\mathbf{\Psi}\right) 
	\nonumber\\
	&+  \beta \,\mbox{tr}\left( \mathbf{\Psi}^\top\sum_{r=1}^S\sum_{s=1}^S\rho_r\rho_s\mathbf{K}_r\mathbf{K}_s\mathbf{\Psi}\mathbf{ L}\right)
	\end{align}
	Since summation and trace operations commute, we have
	\begin{align}
	\xi(\pmb\rho)&
=\sum_{s=1}^S\rho_s\mbox{tr}\left(-2\mathbf{K}_s\pmb\Psi\mathbf{T}^\top\right)+\alpha\mbox{tr}\left(\mathbf{K}_s{\pmb\Psi}\pmb\Psi^\top\right)\nonumber\\
	&+\sum_{r=1}^S\sum_{s=1}^S\rho_r\rho_s\mbox{tr}(\pmb\Psi^\top\mathbf{K}_r\mathbf{K}_s\pmb\Psi(\mathbf{I}_M+\beta\mathbf{L}))\nonumber\\
	&=\sum_{s=1}^S\rho_s\mbox{tr}\left(-2\mathbf{K}_s\pmb\Psi\mathbf{T}^\top\right)+\alpha\mbox{tr}\left(\mathbf{K}_s{\pmb\Psi}\pmb\Psi^\top\right)\nonumber\\
	&+\sum_{r=1}^S\sum_{s=1}^S\rho_r\rho_s\mbox{tr}\left((\mathbf{I}_M+\beta\mathbf{L})^{-\frac{1}{2}}\pmb\Psi^\top\mathbf{K}_r\mathbf{K}_s\pmb\Psi(\mathbf{I}_M+\beta\mathbf{L})^{-\frac{1}{2}}\right)\nonumber\\
	&=\mathbf{b}^\top\pmb\rho+\pmb\rho^\top\mathbf{C}\pmb\rho
	\end{align}
	where $\mathbf{b}\in\mathbb{R}^S$ and $\mathbf{C}\in\mathbb{R}^{S\times S}$ such that the $s$th component of $\mathbf{b}$ is equal to $\mathbf{b}(s)=(-2\mathbf{K}_s\pmb\Psi\mathbf{T}^\top+\alpha\mbox{tr}(\mathbf{K}_s{\pmb\Psi}\pmb\Psi^\top)$ and the $(r,s)$th element of matrix $\mathbf{C}$ is given by
	\begin{align}
	&\mathbf{C}(r,s)\\
	&=\mbox{tr}((\mathbf{I}_M+\beta\mathbf{L})^{-\frac{1}{2}}\pmb\Psi^\top\mathbf{K}_r\mathbf{K}_s\pmb\Psi(\mathbf{I}_M+\beta\mathbf{L})^{-\frac{1}{2}})\nonumber\\
	&=\mathrm{vec}(\mathbf{K}_r\pmb\Psi(\mathbf{I}_M+\beta\mathbf{L})^{-\frac{1}{2}})^\top\mathrm{vec}(\mathbf{K}_s\pmb\Psi(\mathbf{I}_M+\beta\mathbf{L})^{-\frac{1}{2}}),\nonumber
	\end{align}
	using the property that $\mbox{tr}(\mathbf{a}_1^\top\mathbf{a}_2)=\mbox{vec}(\mathbf{a}_1)^\top\mbox{vec}(\mathbf{a}_2)$.
	This shows that $\mathbf{C}$ can be expressed as the Grammian matrix $\mathbf{C}=\mathbf{D}^\top\mathbf{D}$ where
	$\mathbf{D}\in\mathbb{R}^{mn\times S}$ is the matrix whose $s$th column is given by $\mathbf{d}_s=\mathrm{vec}(\mathbf{K}_s\pmb\Psi(\mathbf{I}_M+\beta\mathbf{L})^{-1/2})$. Since Grammian matrices are positive semidefinite \cite{matrixbook}, we have that $\mathbf{C}$ is symmetric and positive semidefinite, and hence $\xi(\pmb\rho)$ is convex. This concludes the proof.
\end{proof}
\begin{proof}[Proposition 2]
	Let $\pmb\rho_2\succeq\pmb\rho_1\succ\mathbf{0}$. For the case when $\pmb\rho\neq\mathbf{0}$, let us  assume $\mathbf{K}\succeq$ is  invertible. This is not an unreasnable requirement as $\mathbf{K}$ is a nonnegative sum ($\pmb\rho\succeq\mathbf{0}$) of kernel matrices: if atleast one of the kernel matrices is positive definite (which is usually the case for most kernel matrices), then $\mathbf{K}$ is positive definite and hence, invertible for $\pmb\rho\neq\mathbf{0}$. Then $\mathbf{B}(\pmb\rho)$ is expressible as
	\begin{equation*}
	\mathbf{B}(\pmb\rho)=-\left[\mathbf{I}_M\otimes (\mathbf{I}_N+\alpha\mathbf{K}^{-1})+\beta\mathbf{L}\otimes \mathbf{I}_N\right]^{-1}
	\end{equation*}
	Since $\pmb\rho_2\succeq\pmb\rho_1$, we have that 
	\begin{eqnarray}
	&&	\sum_{s=1}^S\rho_{1s}\mathbf{K}_s\preceq\sum_{s=1}^S\rho_{2s}\mathbf{K}_s,\,\,\mbox{or}
	\nonumber\\
	&&\left(\sum_{s=1}^S\rho_{1s}\mathbf{K}_s\right)^{-1}\succeq
	\left(\sum_{s=1}^S\rho_{2s}\mathbf{K}_s\right)^{-1},\,\,\mbox{or}\nonumber\\
	&&\mathbf{I}_N+\alpha\left(\sum_{s=1}^S\rho_{1s}\mathbf{K}_s\right)^{-1}\succeq\mathbf{I}_N+\alpha\left(\sum_{s=1}^S\rho_{2s}\mathbf{K}_s\right)^{-1},\,\,\mbox{or}\nonumber\\
	&&\left(\mathbf{I}_M\otimes(\mathbf{I}_N+\alpha\left(\sum_{s=1}^S\rho_{1s}\mathbf{K}_s)^{-1}\right)+\beta\mathbf{L}\otimes\mathbf{I}_N\right)^{-1}\nonumber\\
	&&\preceq\left(\mathbf{I}_M\otimes(\mathbf{I}_N+\alpha\left(\sum_{s=1}^S\rho_{2s}\mathbf{K}_s)^{-1}\right)+\beta\mathbf{L}\otimes\mathbf{I}_N\right)^{-1},\,\,\mbox{or}\nonumber\\
	&&\mathbf{B}(\pmb\rho_1)\succeq\mathbf{B}(\pmb\rho_2)
	\end{eqnarray}
	Since $\pmb\rho=0$ lies in the feasible region, we have that $\mathbf{B}(\pmb\rho)$ takes maximum value at $\pmb\rho=\mathbf{0}$ and the correspondng value is given by setting $\pmb\rho=\mathbf{0}$ in \eqref{B_multi_kernel}, we get that $\mathbf{B}(\mathbf{0})=\mathbf{0}$
	maximum value of $\mbox{vec}(\mathbf{T})^\top \mathbf{B}(\pmb\rho)\mbox{vec}(\mathbf{T})=0$.
\end{proof}

\vfill\pagebreak

\bibliographystyle{IEEEbib}
\bibliography{refs}

\end{document}